\documentclass{article}
\PassOptionsToPackage{numbers}{natbib}  
\usepackage[preprint]{neurips_2026}
\usepackage[utf8]{inputenc}
\usepackage[T1]{fontenc}
\usepackage{hyperref}
\usepackage{url}
\usepackage{booktabs}
\usepackage{amsfonts}
\usepackage{amsmath}
\usepackage{nicefrac}
\usepackage{microtype}
\usepackage{xcolor}
\usepackage{graphicx}
\usepackage{enumitem}
\usepackage{algorithm}
\usepackage{algorithmic}
\DeclareMathOperator*{\argmax}{arg\,max}
\usepackage{multirow}
\usepackage{subcaption}
\usepackage{wrapfig} 
\usepackage[breakable,skins]{tcolorbox}
\newtcolorbox{promptbox}[1][]{%
  enhanced, breakable,
  colback=gray!5, colframe=gray!60,
  fonttitle=\small\bfseries, title={#1},
  left=4pt, right=4pt, top=4pt, bottom=4pt,
  boxrule=0.5pt,
}

\title{APEX: Autonomous Policy Exploration for Self-Evolving LLM Agents}

\author{%
  \textbf{Yibo Li$^{1}$, Jiashuo Yang$^{2}$, Zhi Zheng$^{1}$, Zhiyuan Hu$^{1}$, Yuan Sui$^{1}$,}\\
  \textbf{Shizun Wang$^{1}$, Yufei He$^{1}$, Bryan Hooi$^{1}$}\\
  $^{1}$National University of Singapore\\
  $^{2}$Beijing University of Posts and Telecommunications\\
  {\small \texttt{\{liyibo,dcsbhk\}@nus.edu.sg}}
}

\begin{document}

\maketitle

\begin{abstract}
LLM agents have shown strong performance across a wide range of complex tasks, including interactive environments that require long-horizon decision making. But these agents cannot learn on the fly at test time. Self-evolving agents address this by accumulating memory and reflection across episodes rather than requiring model-weight updates. However, these agents often suffer from \textbf{exploration collapse}: as memory grows, behavior concentrates around familiar high-reward routines, reducing the chance of discovering better alternatives. To address this problem, we propose \textbf{A}utonomous \textbf{P}olicy \textbf{EX}ploration (\textbf{APEX}), which builds and maintains an explicit strategy space through a strategy map---a directed acyclic graph of milestones with prerequisite dependency edges. In APEX, Fork Discovery expands the map with evidence-grounded unexplored directions, while Policy Selection balances exploration and exploitation during planning. Evaluated on nine Jericho text-adventure games and WebArena, a realistic web interaction benchmark, APEX outperforms all baselines. Extensive ablations validate each component's contribution and demonstrate robustness across diverse settings, demonstrating APEX's effectiveness for sustained exploration in self-evolving agents. Code is available at \url{https://github.com/liushiliushi/APEX1}.
\end{abstract}

\section{Introduction}
\label{sec:introduction}

LLM agents can solve many complex tasks, such as software engineering~\cite{swe-bench}, scientific research automation~\cite{aiscientist}, and autonomous computer use~\cite{osbench}. Yet agents that start from scratch each time have a fundamental bottleneck: they do not keep useful experience across runs. As a result, they often repeat similar mistakes and fail to improve. To overcome this limitation, researchers have developed self-evolving agents~\cite{awm, evotest, jitrl}, which systematically internalize experience through memory and reflection to ensure continuous improvement over time.

\begin{figure}[t]
    \centering
    \includegraphics[width=\textwidth]{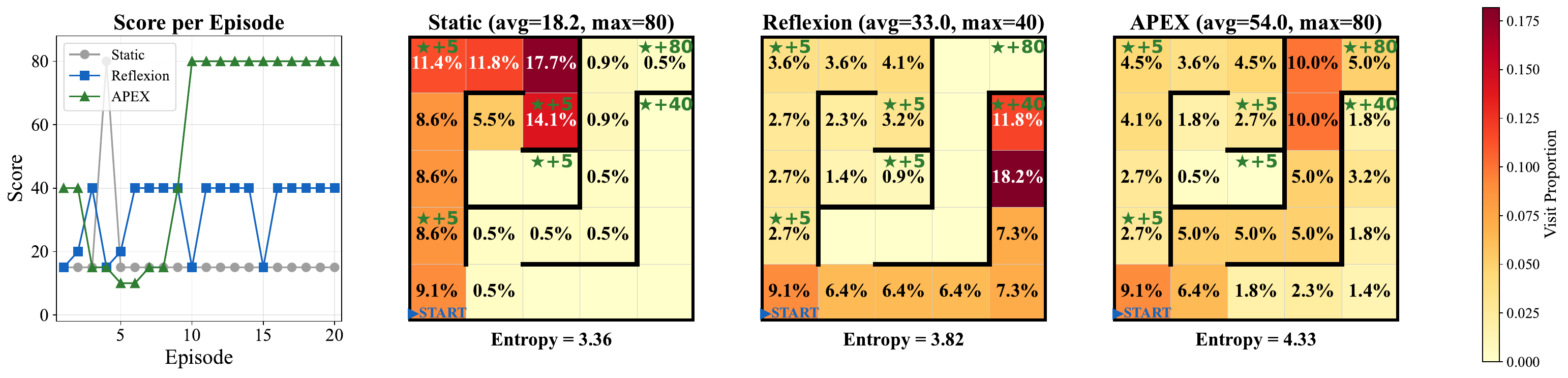}
    \caption{Illustration of exploration collapse in a maze experiment (5$\times$5 grid, 20 episodes, 10 steps each). Room visitation heatmaps (color intensity shows visit proportion; reward cells ($\star$) indicate bonus locations). Static explores broadly but inconsistently. Reflexion locks into a narrow corridor and achieves a higher average while missing high-value rooms. APEX maintains broad coverage and consistently reaches high-reward cells. APEX avoids collapse by explicitly tracking which strategies have been tried and which remain unexplored, and actively directing the agent toward unexplored directions rather than refining familiar ones.}
    \label{fig:teaser}
\end{figure}

Despite these gains, we observe exploration collapse, i.e., agents improve early, then settle into narrow behavioral patterns and stop discovering qualitatively different strategies. Figure~\ref{fig:teaser} illustrates this behavior in a toy maze: a Reflexion agent achieves stable average returns but repeatedly traverses a narrow corridor and misses high-reward regions that require broader exploration. Notably, its maximum score even falls short of a static, non-learning model. Such collapse can be difficult to detect: average returns may remain high even as exploratory diversity decreases, and the agent repeatedly follows a narrow routine, foreclosing the discovery of better alternatives.

The problem is not that agents make bad choices; they never know what choices they are missing. LLM agents face an implicit, unbounded strategy space: unlike RL, there is no predefined action set, and the agent has no natural view of what it has not tried. Discovering unexplored directions requires actively constructing and maintaining an explicit map of what has been tried and what plausible alternatives remain. Two questions follow: (1)~\textbf{how should the strategy space be constructed?} and (2)~\textbf{which strategy should be explored next?} Existing memory and reflection mechanisms provide useful local guidance but do not maintain a persistent, explicit strategy space for decision, making it difficult to sustain broad exploration while still exploiting what has already been learned.

To address this issue, we propose \textbf{APEX} (Autonomous Policy Exploration), a framework centered on an explicit \emph{strategy map}, a directed acyclic graph (DAG) where nodes are milestones and edges encode prerequisite dependencies. The strategy map serves as shared infrastructure for two complementary exploration mechanisms. \textbf{Fork Discovery} drives exploration at the map level: it mines past episodes for directions the agent observed but never pursued, and expands the map's frontier with new milestones. \textbf{Policy Selection} drives exploration at the episode level: it scores milestones via principled uncertainty estimates, ensuring the agent visits under-explored directions rather than repeatedly exploiting familiar ones. Together, they turn exploration into a closed loop---Fork Discovery continuously opens new territory, and Policy Selection ensures the agent visits it.

We evaluate APEX on nine text adventure games from the Jericho benchmark~\cite{jerichogame} and on WebArena~\cite{webarena}, a realistic web interaction benchmark. APEX consistently outperforms all baselines across both settings, with the largest gains on tasks that require discovering qualitatively different strategies---exactly the regime where exploration collapse is most harmful. Ablations and case studies trace these gains to the strategy map: Fork Discovery uncovers directions that reflection-based methods never consider, and Policy Selection ensures they are systematically visited. The results hold robustly across hyperparameter choices and backbone LLMs, underscoring the generality of APEX.

\section{Related Work}
\label{sec:related-work}

\noindent\textbf{Self-Evolving LLM Agents.}
Self-evolving agents improve through accumulated experience across episodes rather than updating model weights. Early work introduced verbal self-reflection~\cite{reflexion}, experience distillation~\cite{awm}, and growing skill libraries~\cite{voyager}. More recent work scales these ideas further. ReasoningBank~\cite{reasoningbank} distills reasoning strategies from both successful and failed experiences into structured memory. EvoTest~\cite{evotest} introduces a gradient-free evolutionary framework that revises an agent's prompt, memory, and tools between episodes. JitRL~\cite{jitrl} adjusts the action logits of the current episode by retrieving relevant historical trajectories from memory, enabling continual improvement without gradient updates. However, none of these systems builds an explicit strategy space to guide exploration; this is the gap APEX addresses.

\noindent\textbf{Exploration in Fixed Action Spaces.}
Classical exploration methods assume a predefined primitive action space. They fall into two broad families: \textit{intrinsic motivation}, which rewards agents for visiting novel states or exhibiting diverse behaviors, including count-based bonuses~\cite{count1, count2}, curiosity-driven rewards~\cite{curiosity1, curiosity2}, episodic curiosity~\cite{episodiccuriosity}, and diversity-maximizing skill discovery~\cite{diversity}; and \textit{planning under uncertainty}, which models the agent's uncertainty to guide exploration, including UCB~\cite{UCB, MCTS}, Thompson Sampling~\cite{thompson1,thompson2}, and maximum entropy objectives~\cite{entropy}. Despite their variety, all share a fundamental assumption: the primitive action space is fixed and given. They answer the question of which known option to try next, but never ask what options should exist in the first place. APEX addresses this complementary question: Fork Discovery actively constructs the strategy space from trajectory evidence, while Policy Selection operates over it.

\noindent\textbf{Entropy Bottleneck in LLM Reinforcement Learning.}
Recent work identifies entropy collapse in LLM Reinforcement Learning (RL). \citet{entropy1} show that RL training systematically reduces policy entropy, causing the agent to concentrate probability mass on familiar high-reward actions and progressively lose behavioral diversity. \citet{entropy2} show that RLVR improves average performance but narrows the reasoning boundary, as RL sharpens the distribution over existing paths rather than discovering new ones. These works focus on the weight-update regime, while APEX addresses the analogous collapse in test-time self-evolving agents.
\vspace{-2mm}
\section{Exploration Collapse: An Empirical Study}
\label{sec:collapse}

We study how Reflexion~\cite{reflexion}, a representative self-evolving agent, and a memoryless Static baseline perform across multiple tasks: a 5$\times$5 maze with 20 episodes of 10 steps each, and three text adventure games, Zork3, Deephome, and Ztuu, with 50 episodes of 120 steps each. Both use Gemini 3 Flash as the backbone LLM with identical settings. In the maze, the agent receives a localized text observation describing the current room and its available exits, without global coordinates. The action space consists of the cardinal directions reachable from the current cell.

\begin{figure}[t]
    \centering
    \includegraphics[width=\textwidth]{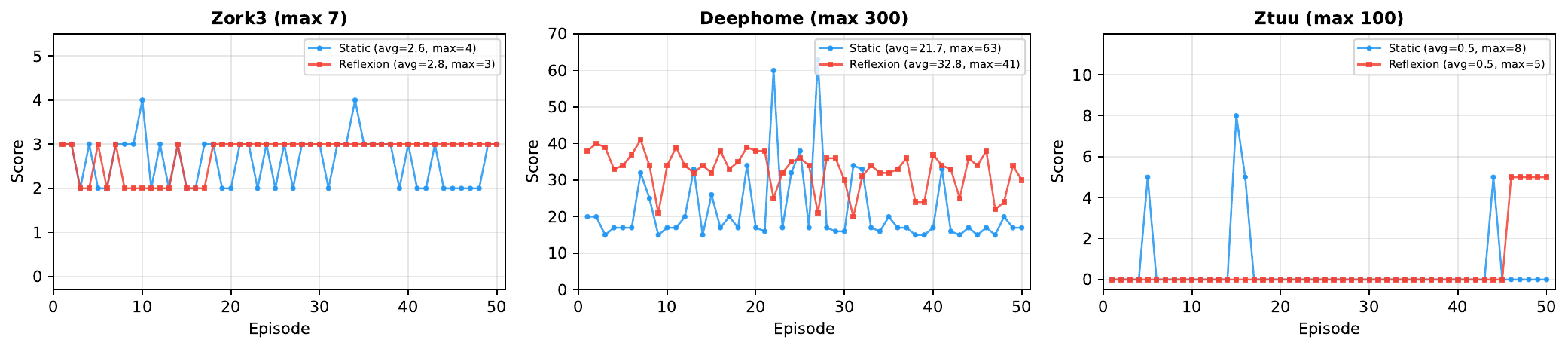}
    \caption{Comparison of Static and Reflexion on three Jericho games. Reflexion often achieves a higher average but a lower maximum than Static, indicating convergence to a narrow behavioral distribution that misses superior strategies.}
    \label{fig:collapse}
    \vspace{-3mm}
    \vspace{-3mm}
\end{figure}

As shown in Figure~\ref{fig:teaser},Reflexion converges to a fixed route that reliably reaches the +40 reward but never discovers the +80 cell, while Static occasionally reaches the +80 cell. The heatmap of Reflexion shows heavy concentration in that corridor. APEX, by contrast, explores the maze broadly and consistently reaches the high-reward cell. As shown in Figure~\ref{fig:collapse}, the same pattern holds across three text adventure games of varying difficulty: in most cases, Reflexion achieves a higher average but a lower maximum than Static.

Once Reflexion finds a rewarding strategy, the agent repeats it episode after episode and stops exploring different approaches. Trapped in a local optimum, it never discovers the superior strategies that Static occasionally stumbles upon through sheer behavioral diversity. This is \textbf{exploration collapse}: the agent converges to a narrow behavioral routine and stops searching for better alternatives.

\vspace{-2mm}
\section{Method: APEX}
\label{sec:method}

\begin{figure}[t]
    \centering
    \includegraphics[width=\textwidth]{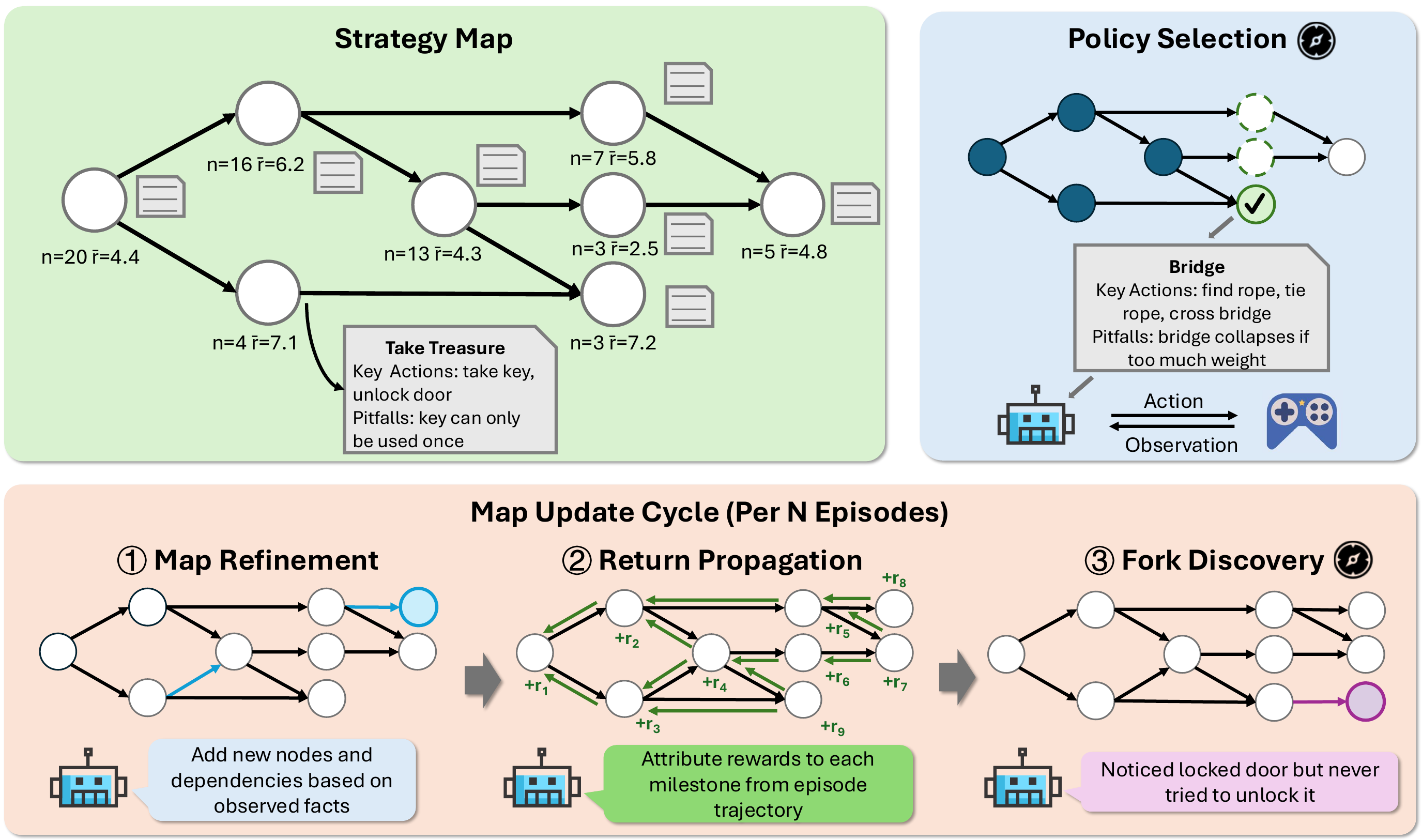}
    \caption{Overview of APEX. Each node in the strategy map displays $n$ (visit count) and $\bar{r}$ (average reward). Within each episode, \textbf{Policy Selection} is invoked repeatedly to choose the next milestone. When a milestone is achieved or failed, Policy Selection is called again until the episode ends. Every $N$ episodes, a periodic reflection cycle runs \textbf{Map Refinement} to update existing nodes and edges, \textbf{Return Propagation} to update node statistics on the refined map, and \textbf{Fork Discovery} to expand the map with new milestones and dependencies.}
    \label{fig:method}
\end{figure}

\noindent\textbf{Overview.} As shown in Figure ~\ref{fig:method}, APEX addresses exploration collapse by maintaining an explicit \emph{strategy map} that records what has been tried and what plausible alternatives remain. For example, if prior trajectories have identified three milestones (e.g., three doors), the map should direct the agent toward discovering novel alternatives (e.g., a fourth door), as well as revisiting under-explored ones.

\noindent\textbf{Strategy Map.}
The strategy map $\mathcal{G} = (V, E, \theta)$ is a Directed Acyclic Graph (DAG) where each node $v \in V$ is a milestone (a discrete sub-goal described in natural language with associated key actions, e.g., ``collect the brass lantern'' via go north, take lantern), and each edge $(u, v) \in E$ encodes that $u$ must be achieved before $v$. Each node carries statistics $\theta_v = (n(v),\, \bar{r}(v),\, \text{Var}(v))$: visit count, average reward, and reward variance. A designated root node $v_0 \in V$ represents the initial state. Milestones with no prerequisites have $v_0$ as their sole dependency.

The map defines a higher-level planning problem. The abstract state $s \subseteq V$ is the set of milestones achieved so far in the current episode. We denote by $\mathcal{A}(s)$ the set of eligible milestones---those not yet achieved whose predecessors are all in $s$:
\begin{equation}
    \mathcal{A}(s) = \bigl\{v \notin s : \{u \mid (u,v)\in E\} \subseteq s\bigr\}.
\end{equation}
We formalize this as a Milestone Abstract MDP $\mathcal{M} = (\mathcal{S}, \mathcal{A}, \mathcal{T}, R, \gamma)$, where $\mathcal{S}$ is the set of all valid abstract states, $\mathcal{A}(s)$ is the eligible milestone set defined above, $\mathcal{T}(s \cup \{v\} \mid s, v)$ is the transition probability from $s$ to $s \cup \{v\}$ upon attempting milestone $v$, and $R(s, v)$ is the distribution of rewards upon attempting $v$,  which we approximate using $\bar{r}(v)$.

\subsection{Policy Selection}
\label{sec:selection}

Given the current state $s$, Policy Selection picks the next milestone to pursue from the eligible set $\mathcal{A}(s)$, reducing the open-ended action space to a finite bandit problem over a small set of DAG-structured candidates. Each eligible milestone $v \in \mathcal{A}(s)$ is scored by a function $f$ of its bandit statistics, and the highest-scoring one is selected:
\begin{equation}
    \text{score}(v) = f\bigl(n(v),\, \bar{r}(v),\, \text{Var}(v)\bigr), \qquad v^* = \argmax_{v \in \mathcal{A}(s)}\, \text{score}(v).
\end{equation}
The choice of $f$ determines the exploration-exploitation tradeoff. Three bandit instantiations are:
\begin{itemize}[leftmargin=*]
\item \textbf{Thompson Sampling}~\cite{thompson1,thompson2}: $\text{score}(v) = \text{sample}\!\left(\mathcal{N}\!\left(\bar{r}(v),\, \sqrt{\text{Var}(v)/n(v)}\right)\right)$. High-variance milestones remain under exploration even when their average reward is low.
\item \textbf{UCB}~\cite{UCB}: $\text{score}(v) = \bar{r}(v) + c\sqrt{\log \!\bigl(\sum_{u \in \mathcal{A}(s)} n(u)\bigr) / n(v)}$, where $\sum_{u \in \mathcal{A}(s)} n(u)$ is the total visit count over all currently eligible milestones. The bonus decreases as a milestone is visited more, naturally shifting focus toward less-explored candidates.
\item \textbf{$\epsilon$-Greedy}~\cite{e-greedy}: selects the highest $\bar{r}(v)$ milestone with probability $1-\epsilon$, and a random eligible milestone with probability $\epsilon$.
\end{itemize}
We use Thompson Sampling as the default, as it requires no hyperparameter tuning and is naturally variance-aware. Unvisited milestones ($n(v) = 0$) are assigned $+\infty$ score and always selected first, with statistics initialized after the first visit via Return Propagation (Section~\ref{sec:backprop}).

Policy Selection proceeds sequentially during episode execution: after each milestone is achieved or fails, we score all milestones whose dependencies have been actually achieved, select the highest-scoring eligible one, and execute based on the milestone. This is more robust than fixing a full milestone sequence upfront: if the plan were fixed and a milestone fails, the agent would still attempt its dependents even though their prerequisites were never met. With sequential selection, failed milestones never satisfy their dependents' prerequisites, so those dependents never enter $\mathcal{A}(s)$, and the agent naturally continues with other reachable branches.

\subsection{Map Refinement}
\label{sec:refinement}

After each episode $t$, an LLM generates a structured summary $\tilde{h}_t$ from the raw trajectory $h_t$:
\begin{equation}
    \tilde{h}_t = \mathrm{LLM}_{\text{sum}}(h_t).
\end{equation} Every $N$ episodes, the reflection cycle operates on $\tilde{H}_t = \{\tilde{h}_{t-N+1}, \ldots, \tilde{h}_t\}$, the summaries from the most recent $N$ episodes. As the first step of the reflection cycle, Map Refinement updates existing nodes based on observed facts from recent episodes. It refines key actions when better sequences are discovered, corrects dependency edges when assumed prerequisites turn out to be wrong, and prunes duplicate nodes. Formally, the LLM receives $\mathcal{G}_t$ and $\tilde{H}_t$ and returns:
\begin{equation}
    (\Delta V^M,\, \Delta E^M) = \mathrm{LLM}_{\text{map}}(\mathcal{G}_t,\, \tilde{H}_t),
\end{equation}
where $\Delta V^M$ is the set of node changes, including fact-grounded additions, attribute updates such as refined key actions, and deduplication pruning, and $\Delta E^M$ is the corrected edge set. These are applied to $\mathcal{G}_t$ to produce the refined map $\mathcal{G}'_t = (V_t \cup \Delta V^M,\; E_t \cup \Delta E^M,\; \theta_t)$. The key distinction from Fork Discovery is intent: Map Refinement encodes observed facts and corrects errors, while Fork Discovery adds nodes grounded in unexplored possibilities.

\subsection{Return Propagation}
\label{sec:backprop}

Every $N$ episodes, after Map Refinement, Return Propagation updates node statistics in $\mathcal{G}'_t$ using $\tilde{H}_t = \{\tilde{h}_{t-N+1}, \ldots, \tilde{h}_t\}$, the summaries of the most recent $N$ episodes. For each episode summary $\tilde{h}_i \in \tilde{H}_t$, an LLM attributes a scalar reward $r(v)$ to each milestone $v$ attempted in that episode based on the observed score gains:
\begin{equation}
    \{r(v)\}_{v \in V_a} = \mathrm{LLM}_{\text{reward}}(\tilde{h}_i, \mathcal{G}'_t).
\end{equation}
Let $V_a$ denote the set of milestones attempted in episode $i$. For each $v \in V_a$, we compute a credit score recursively along the DAG edges:
\begin{equation}
    G_v = r(v) + \gamma \sum_{u \in \mathrm{succ}(v) \cap V_a} G_u,
\end{equation}
where $\mathrm{succ}(v)$ denotes the direct successors of $v$ in $\mathcal{G}$, and $\gamma \in [0, 1]$ controls how much downstream credit propagates back. This recursion is computed via a reverse topological pass, ensuring that $v$ only receives credit from milestones it directly enables, not from parallel branches. For each $v \in V_a$, $n(v)$ is incremented and $\bar{r}(v)$ is updated as a running mean over $G_v$ values. Milestones that were never scheduled in the episode are skipped and their statistics remain unchanged. This gives credit to enabling milestones that yield little reward on their own but unlock high-reward successors (e.g., ``obtain the key'' enables ``unlock the treasure room'').

\subsection{Fork Discovery}
\label{sec:dpm}

Return Propagation refines what the agent knows about existing milestones, but it cannot discover what it has never tried. As the agent accumulates experience, the map must grow to cover new territory. To actively push the map's frontier outward, we introduce \textbf{Fork Discovery}, which generates new milestones by combining the current map with information observed during gameplay, such as available actions and unexplored directions that the agent has yet to pursue.

Every $N$ episodes, an LLM receives $\mathcal{G}'_t$ and $\tilde{H}_t$, identifies actions the agent observed or had available but never pursued, and proposes new milestones with dependency edges:
\begin{equation}
    (\Delta V^F,\, \Delta E^F) = \mathrm{LLM}_{\text{fork}}(\mathcal{G}'_t,\, \tilde{H}_t).
\end{equation}
After all three steps, the updated map is:
\begin{equation}
    \mathcal{G}_{t+1} = (V_t \cup \Delta V^M \cup \Delta V^F,\; E_t \cup \Delta E^M \cup \Delta E^F,\; \theta_{t+1}),
\end{equation}
where each new node $v \in \Delta V^F$ is initialized with $n(v) = 0$.

Together, the four components form a coherent system: Fork Discovery and Policy Selection drive exploration: the former expands the map's frontier and the latter directs the agent within it. Map Refinement and Return Propagation keep the map accurate and its statistics current. As the strategy map matures, Fork Discovery is disabled after a configurable episode threshold, naturally shifting the system from exploration toward exploitation. The complete pseudocode is given in Appendix~\ref{sec:workflow}. See Appendix~\ref{sec:impl-details} for implementation details and prompt templates.

\section{Experiments}
\label{sec:experiments}

We design our experiments to answer the following four research questions:
\begin{itemize}[leftmargin=*]
    \item \textbf{RQ1}: How does APEX perform compared to state-of-the-art baselines?
    \item \textbf{RQ2}: How do the strategy map structure, Fork Discovery, and Policy Selection each contribute to APEX's performance?
    \item \textbf{RQ3}: What does APEX actually discover, and why does it outperform baselines? 
    \item \textbf{RQ4}: How sensitive is APEX to key hyperparameters such as the reflection interval and the choice of backbone LLM?
\end{itemize}

\subsection{Setup}

\noindent\textbf{Benchmarks.} We evaluate on two environments to test APEX across different task types:
\begin{itemize}[leftmargin=*]
    \item \textbf{Jericho}~\cite{jerichogame}: A benchmark suite for interactive fiction games where agents interact purely via textual commands. We evaluate on nine games spanning diverse challenges: Zork1, Balances, Zork3, Temple, Pentari, Ludicorp, Deephome, Detective, and Ztuu.
    \item \textbf{WebArena}~\cite{webarena}: A realistic web environment comprising multiple functional websites. It challenges agents to navigate complex interfaces and execute sequential actions to fulfill user instructions. We evaluate on all 812 tasks across five website categories; each task is attempted for 5 episodes.
\end{itemize}

\noindent\textbf{Metric.} We report \textbf{Final-$K$}: the mean performance over the final $K$ episodes. For Jericho, $K{=}5$ over 50 episodes per game. For WebArena, $K{=}3$ over 5 episodes per task.

\noindent\textbf{Baselines.} We evaluate five baselines: (1)~\textbf{Static}: a non-learning agent with no cross-episode memory. (2)~\textbf{Memory}: an in-context learning baseline that accumulates past episode transcripts into the agent's context window, using a FIFO strategy to handle token limits. (3)~\textbf{Reflexion}~\cite{reflexion}: a prompt-based method where the agent generates structured textual self-reflections after each episode to guide subsequent attempts. (4)~\textbf{EvoTest}~\cite{evotest}: an evolutionary framework that iteratively revises the agent's prompt, memory, and tools between episodes, but without maintaining an explicit strategy map. (5)~\textbf{ACE}~\cite{ACE}: a framework that treats contexts as evolving playbooks, accumulating and refining strategies through a modular grow-and-refine process, without maintaining an explicit strategy map. All methods, including APEX, use Gemini 3 Flash as the backbone LLM with identical settings. Detailed baseline descriptions are in Appendix~\ref{sec:baselines}.

\subsection{Main Results}

We address \textbf{RQ1} by comparing APEX against all baselines across Jericho and WebArena.

\begin{table}[t]
\centering
\caption{Results on Jericho games. Final-5: mean score of episodes 46--50.}
\label{tab:cross-game-results}
\resizebox{\textwidth}{!}{%
\begin{tabular}{@{}l r r r r r r r r r@{}}
\toprule
\textbf{Method} & \textbf{Zork1} & \textbf{Balances} & \textbf{Zork3} & \textbf{Temple} & \textbf{Pentari} & \textbf{Ludicorp} & \textbf{Deephome} & \textbf{Detective} & \textbf{Ztuu} \\
\midrule
Static      & 39.8 & 11.0 & 2.8  & 6.2  & 6.0  & 13.6 & 17.2 & 70.0  & 0.0  \\
Memory      & 46.0 & 13.0 & 2.4 & 6.2  & 7.0  & 17.0 & 29.0 & 130.0 & 0.0  \\
Reflexion   & 59.8 & 20.0 & 3.0 & 9.0  & 10.0 & 35.0 & 29.6 & 144.0 & 5.0  \\
EvoTest     & 42.0 & 18.0 & 2.4 & 10.0 & 16.0 & 48.0 & 56.8 & 314.0 & 25.0 \\
ACE         & 46.0 & 18.0 & 2.2 & 7.2  & 19.0 & 19.2 & 8.4  & 198.0 & 3.4  \\
APEX (Ours) & \textbf{73.0} & \textbf{23.0} & \textbf{4.6} & \textbf{20.8} & \textbf{46.0} & \textbf{58.8} & \textbf{103.4} & \textbf{322.0} & \textbf{29.0} \\
\bottomrule
\end{tabular}%
}
\end{table}

\begin{table}[t]
\centering
\setlength{\tabcolsep}{4pt}
\begin{minipage}[t]{0.49\textwidth}
\centering
\caption{Results on WebArena (Final-3)}
\label{tab:webarena-results-avg}
\small
\setlength{\tabcolsep}{3pt}
\begin{tabular}{@{}l@{\hspace{2pt}}rrrrr@{}}
\toprule
\textbf{Method} & \textbf{Shopping} & \textbf{Reddit} & \textbf{GitLab} & \textbf{CMS} & \textbf{Map} \\
\midrule
Static      & 34.7 & 45.7 & 36.3 & 44.8 & 56.3 \\
Memory      & 39.1 & 53.8 & 53.9 & 53.2 & 55.2 \\
Reflexion   & 34.1 & 36.2 & 50.8 & 46.7 & 54.2 \\
EvoTest     & 38.9 & 50.9 & 54.4 & 53.8 & 52.9 \\
ACE         & 35.9 & 52.5 & 51.6 & 52.4 & 55.7 \\
APEX        & \textbf{42.9} & \textbf{68.2} & \textbf{60.2} & \textbf{55.1} & \textbf{57.6} \\
\bottomrule
\end{tabular}
\end{minipage}
\hfill
\begin{minipage}[t]{0.49\textwidth}
\centering
\caption{Cross-task generalization on WebArena.}
\label{tab:webarena-run1}
\small
\setlength{\tabcolsep}{3pt}
\begin{tabular}{@{}l@{\hspace{2pt}}rrrrr@{}}
\toprule
\textbf{Method} & \textbf{Shopping} & \textbf{Reddit} & \textbf{GitLab} & \textbf{CMS} & \textbf{Map} \\
\midrule
Static      & 33.9 & 44.5 & 35.9 & 43.9 & 49.2 \\
Memory      & 37.5 & 52.7 & 53.2 & \textbf{53.1} & 55.1 \\
Reflexion   & 32.8 & 34.9 & 49.5 & 46.2 & 48.6 \\
EvoTest     & 37.5 & 47.7 & 53.4 & 52.7 & 50.4 \\
ACE         & 26.6 & 46.5 & 50.5 & 48.9 & 51.4 \\
\textbf{APEX}        & \textbf{41.7} & \textbf{56.6} & \textbf{55.9} & 52.7 & \textbf{56.0} \\
\bottomrule
\end{tabular}
\end{minipage}

\end{table}

\begin{figure}[t]
  \centering
  \includegraphics[width=\linewidth]{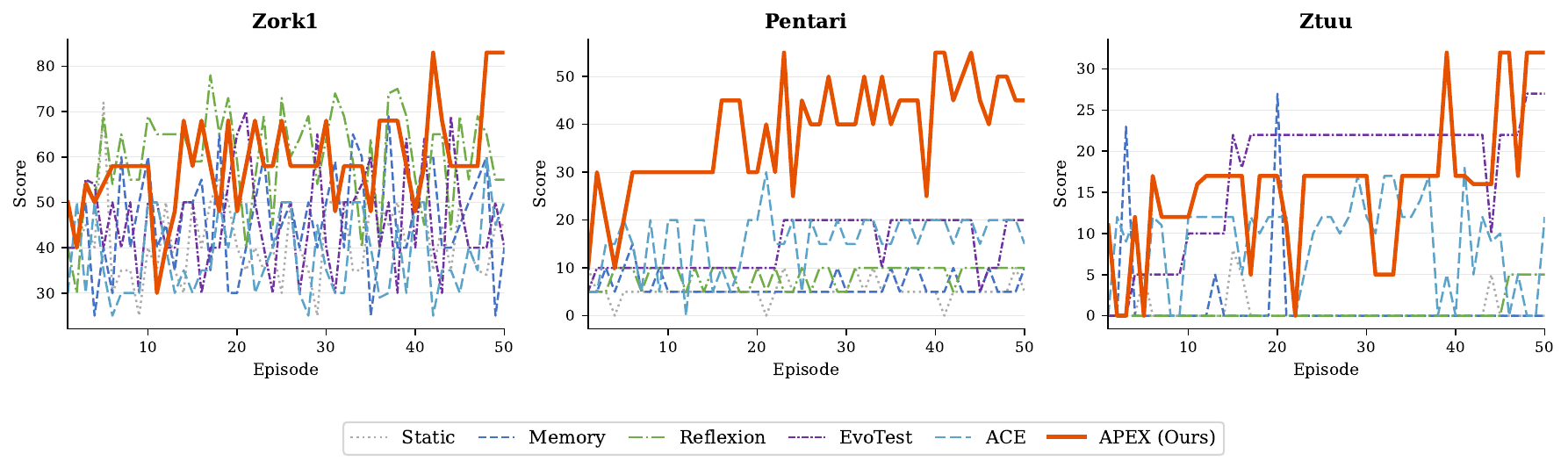}
  \caption{Per-episode scores on three representative Jericho games.}
  \label{fig:learning-curves}
\end{figure}

\noindent\textbf{Jericho Results.} As shown in Table~\ref{tab:cross-game-results}, APEX achieves the best Final-5 on all nine games. Figure~\ref{fig:learning-curves} shows per-episode learning curves on three representative games; full curves for all nine games are in Appendix~\ref{sec:additional-results}. The gains are most pronounced in games that require discovering qualitatively different strategies across episodes, where the strategy map enables systematic exploration beyond what reflection or memory alone can achieve. APEX outperforms all baselines, showing that structured exploration is beneficial even in games with strong single-path strategies. Overall, the results suggest that structured strategy spaces are most valuable when the task rewards diversity of exploration rather than depth along a single route.

\noindent\textbf{WebArena Results.}
We further evaluate on WebArena, where agents interact with structured web pages. As shown in Table~\ref{tab:webarena-results-avg}, APEX achieves the best Final-3 score and leads on all five categories, outperforming the next best baseline by a clear margin. The gains are most pronounced on Reddit and GitLab, where tasks require multi-step sequences with non-obvious intermediate actions---precisely the setting where Fork Discovery adds value by encoding untried action paths into the strategy map. Overall, APEX's structured exploration mechanism transfers effectively beyond text adventure games to realistic web environments.

\noindent\textbf{Cross-Task Generalization.}
Table~\ref{tab:webarena-run1} evaluates cross-task generalization on WebArena, where each task is attempted without any task-specific prior experience. Performance therefore reflects only what the agent has learned from other tasks. APEX leads across the majority of domains and achieves the best overall score, showing that the structured knowledge encoded in the strategy map transfers effectively to unseen tasks rather than merely memorizing task-specific solutions.

\vspace{-2mm}
\subsection{Ablation Studies}
\label{sec:ablation}

We address \textbf{RQ2} by isolating each design choice (the map structure, space expansion method, and selection algorithm), varying one at a time while keeping the rest fixed at the default configuration: DAG strategy map, Thompson Sampling, Fork Discovery, and reflect interval $N{=}5$.

\noindent\textbf{Strategy Space Representation: DAG vs.\ Flat List.}
We compare two representations: (1)~\textbf{Flat List}, milestones maintained as an unordered list without dependency edges, and (2)~\textbf{DAG (Ours)}, milestones connected by multi-prerequisite dependency edges. As shown in Table~\ref{tab:representation-ablation}, DAG achieves the best Final-5 on all three games. Flat List underperforms DAG across all games: without dependency structure, the agent cannot reason about milestone ordering or reachability, limiting effective exploration. The results confirm that capturing multi-prerequisite dependencies is key to the strategy map's effectiveness.

\begin{table}[t]
\centering
\vspace{-4mm}
\begin{minipage}[t]{0.48\textwidth}
\centering
\caption{Strategy space representation comparison (Final-5).}
\label{tab:representation-ablation}
\small
\begin{tabular}{@{}lrrr@{}}
\toprule
\textbf{Representation} & \textbf{Zork1} & \textbf{Pentari} & \textbf{Ztuu} \\
\midrule
Flat List          & 52.1 & 27.0 & 12.0 \\
\textbf{DAG (Ours)} & \textbf{73.0} & \textbf{46.0} & \textbf{29.0} \\
\bottomrule
\end{tabular}
\end{minipage}%
\hfill
\begin{minipage}[t]{0.48\textwidth}
\centering
\caption{Fork Discovery ablation (Final-5).}
\label{tab:discovery-ablation}
\small
\begin{tabular}{@{}lrrr@{}}
\toprule
\textbf{Method} & \textbf{Zork1} & \textbf{Pentari} & \textbf{Ztuu} \\
\midrule
No FD           & 65.0 & 24.0 & 5.0 \\
\textbf{FD (Ours)} & \textbf{73.0} & \textbf{46.0} & \textbf{29.0} \\
\bottomrule
\end{tabular}
\end{minipage}
\end{table}

\begin{figure}[!ht]
    \centering
    \begin{subfigure}[t]{0.235\textwidth}
        \centering
        \includegraphics[width=\linewidth]{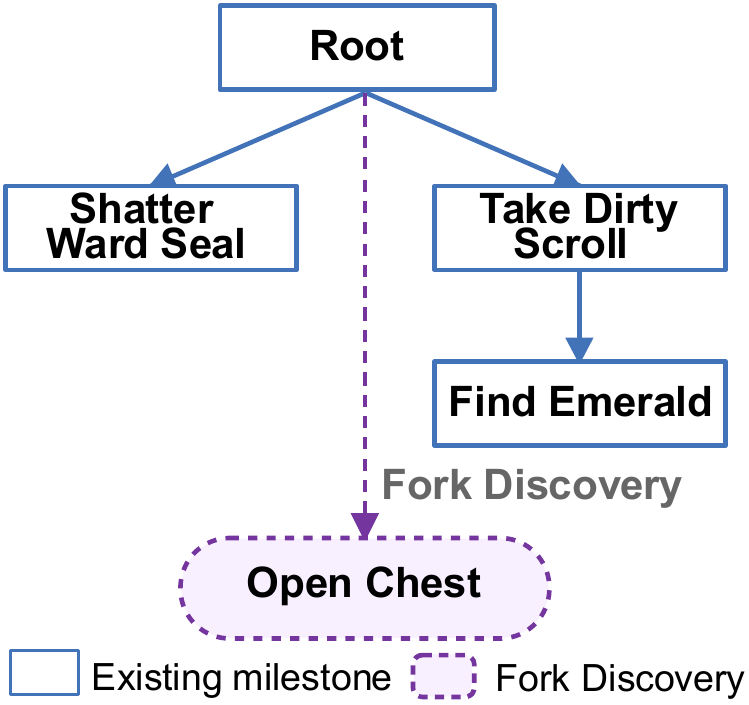}
        \caption{Pentari: Map}
    \end{subfigure}\hfill
    \begin{subfigure}[t]{0.235\textwidth}
        \centering
        \includegraphics[width=\linewidth]{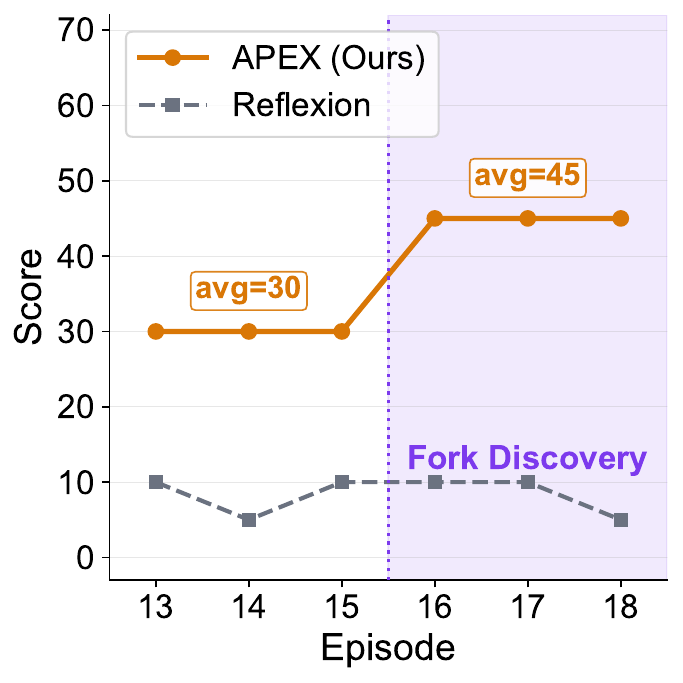}
        \caption{Pentari: Score}
    \end{subfigure}\hfill
    \begin{subfigure}[t]{0.235\textwidth}
        \centering
        \includegraphics[width=\linewidth]{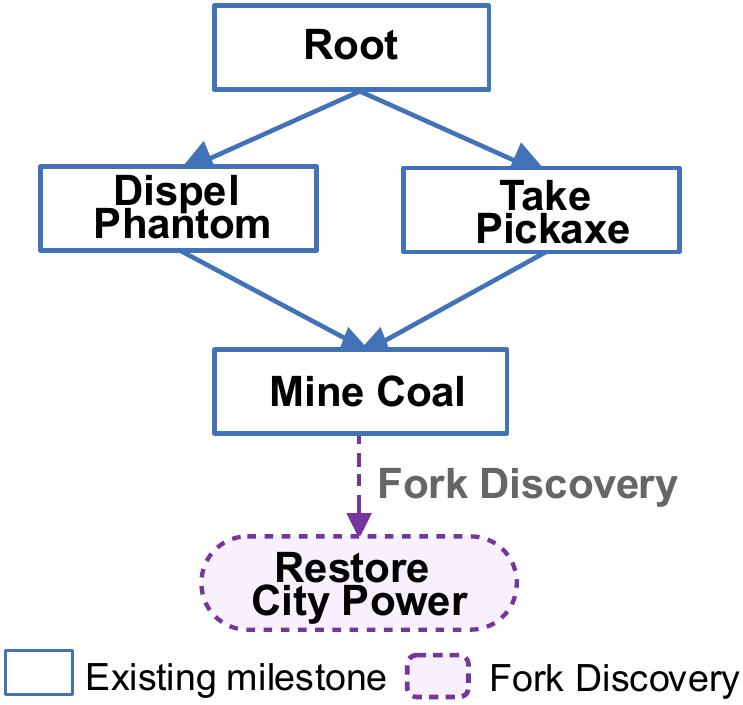}
        \caption{Deephome: Map}
    \end{subfigure}\hfill
    \begin{subfigure}[t]{0.235\textwidth}
        \centering
        \includegraphics[width=\linewidth]{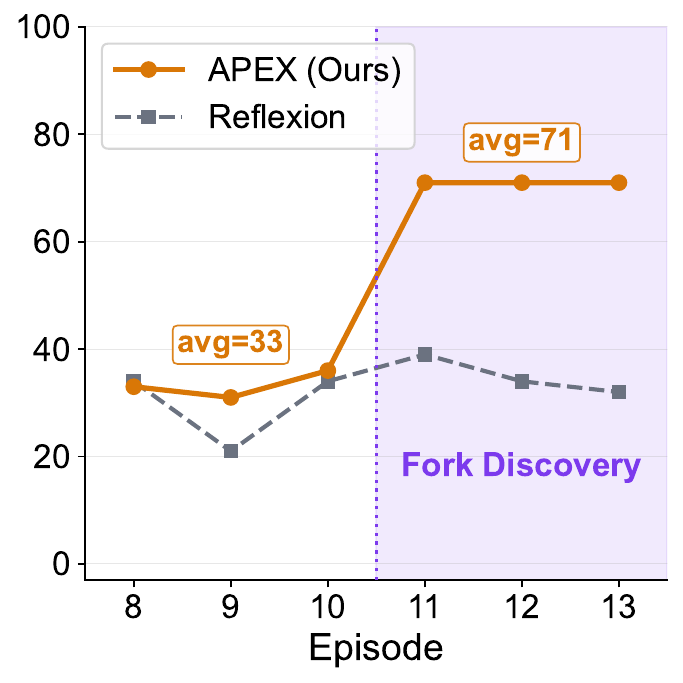}
        \caption{Deephome: Score}
    \end{subfigure}
    \caption{Fork Discovery case studies on Pentari (a,b) and Deephome (c,d): strategy maps and per-episode scores before and after Fork Discovery adds a new milestone.}
    \vspace{-3mm}
    \label{fig:case-study}
\end{figure}

\begin{table}[t]
\centering
\begin{minipage}[t]{0.48\textwidth}
\centering
\caption{Policy Selection ablation (Final-5).}
\label{tab:selection-ablation}
\small
\begin{tabular}{@{}lrrr@{}}
\toprule
\textbf{Method} & \textbf{Zork1} & \textbf{Pentari} & \textbf{Ztuu} \\
\midrule
$\epsilon$-Greedy & 48.0 & 35.0 & 1.0 \\
UCB             & 59.7 & 43.0 & 12.0 \\
\textbf{Thompson (Ours)} & \textbf{73.0} & \textbf{46.0} & \textbf{29.0} \\
\bottomrule
\end{tabular}
\end{minipage}%
\hfill
\begin{minipage}[t]{0.48\textwidth}
\centering
\caption{Reflect interval $N$ sensitivity (Final-5).}
\label{tab:reflect-interval}
\small
\begin{tabular}{@{}lrrr@{}}
\toprule
\textbf{$N$} & \textbf{Zork1} & \textbf{Pentari} & \textbf{Ztuu} \\
\midrule
$1$  & 34.6 & 16.0 & 25.0 \\
$2$  & \textbf{73.0} & 26.0 & 24.0 \\
$5$  & 70.2 & \textbf{46.0} & \textbf{29.0} \\
$10$ & 52.8 & 30.0 & 22.0 \\
\bottomrule
\end{tabular}
\end{minipage}
\end{table}

\noindent\textbf{Fork Discovery: FD vs.\ No FD.}
The strategy map must grow as the agent accumulates experience, but whether to explore matters. We compare two approaches: No FD updates the map only with milestones the agent has directly visited, reinforcing known paths without identifying unexplored directions; Fork Discovery, by contrast, mines past episodes for alternatives the agent observed but never pursued, actively expanding the map's frontier. As shown in Table~\ref{tab:discovery-ablation}, without Fork Discovery, the agent can only reinforce what it has already tried; the gap is smaller on Zork1 but large on Ztuu, where higher scores require discovering directions the agent has never pursued. Fork Discovery closes this gap by actively expanding the map's frontier as the agent gains experience.

\noindent\textbf{Policy Selection: Thompson vs.\ UCB vs.\ $\epsilon$-Greedy.}
We compare three selection strategies: $\epsilon$-Greedy ignores uncertainty and selects randomly with probability $\epsilon$; UCB adds a count-based optimism bonus toward less-visited milestones; Thompson Sampling draws from each milestone's posterior, naturally favoring high-uncertainty ones. As shown in Table~\ref{tab:selection-ablation}, both UCB and Thompson outperform $\epsilon$-Greedy, confirming that leveraging uncertainty is more effective than random exploration. Thompson outperforms UCB across all three games, with the largest gap on Ztuu where variance-awareness provides the most benefit.

\noindent\textbf{Return Propagation: DAG vs.\ Sequential.}
We compare two credit assignment schemes for Return Propagation: \textbf{Sequential} discounts rewards along the episode path order: given the ordered sequence $(v_1, \ldots, v_k)$ of attempted milestones, $G_{v_i} = r(v_i) + \gamma \cdot G_{v_{i+1}}$, so each milestone inherits credit from all later milestones regardless of dependency structure. \textbf{DAG} instead propagates credit strictly along prerequisite edges: $G_v = r(v) + \gamma \sum_{u \in \mathrm{succ}(v) \cap V_a} G_u$, ensuring a milestone only receives credit from milestones it directly enables. As shown in Table~\ref{tab:backprop-ablation}, DAG credit assignment outperforms Sequential across all three games. Sequential conflates credit across parallel branches, assigning reward to milestones that did not causally enable it. DAG restricts credit flow to actual prerequisite edges, producing more accurate value estimates and better-informed Policy Selection.

\begin{table}[t]
\vspace{-4mm}
\begin{minipage}[t]{0.4\linewidth}
\centering
\vbox to 1.4\baselineskip{%
  \captionof{table}{Return Propagation ablation.}%
  \label{tab:backprop-ablation}%
  \vfil}
\small
\begin{tabular}{@{}lrrr@{}}
\toprule
\textbf{Method} & \textbf{Zork1} & \textbf{Pentari} & \textbf{Ztuu} \\
\midrule
Sequential          & 71.6 & 35.0 & 17.0 \\
\textbf{DAG (Ours)} & \textbf{73.0} & \textbf{46.0} & \textbf{29.0} \\
\bottomrule
\end{tabular}
\end{minipage}%
\hfill%
\begin{minipage}[t]{0.58\linewidth}
\centering
\vbox to 1.4\baselineskip{%
  \captionof{table}{Comparison of cost (\$) among all the methods.}%
  \label{tab:cost}%
  \vfil}
\small
\setlength{\tabcolsep}{4pt}
\begin{tabular}{@{}lrrrrrr@{}}
\toprule
 & \textbf{Static} & \textbf{Memory} & \textbf{Reflexion} & \textbf{EvoTest} & \textbf{ACE} & \textbf{APEX} \\
\midrule
Cost & 278 & 320 & 306 & 306 & 324 & 318 \\
\bottomrule
\end{tabular}
\end{minipage}
\vspace{-2mm}
\end{table}

\subsection{Analysis: What Does APEX Discover?}
\label{sec:analysis}

We address \textbf{RQ3} by inspecting the strategy maps APEX builds in practice. Figure~\ref{fig:case-study} shows two representative Fork Discovery events and their score impact.

In \textbf{Pentari}, Fork Discovery identifies ``Open  Chest'', a milestone the agent had encountered but never deliberately attempted, and adds it directly to the strategy map. Scores jump from an average of 28 to 43, a level Reflexion never reaches. In \textbf{Deephome}, Fork Discovery encodes ``Restore City Power'', which requires both ``Dispel Phantom'' and ``Take Pickaxe'' via ``Mine Coal''. Scores jump from 33 to 71, while Reflexion plateaus at 30--40 throughout. Both cases show that Fork Discovery surfaces valuable directions the agent had evidence for but never explicitly pursued.

\subsection{Sensitivity Analysis}
\label{sec:sensitivity}

We address \textbf{RQ4} by examining how APEX's performance varies with the reflection interval and the choice of backbone LLM.

\noindent\textbf{Reflect Interval.}
Fork Discovery and Map Refinement run periodically every $N$ episodes. A smaller $N$ means more frequent map updates but higher LLM cost, while a larger $N$ means less overhead but slower adaptation. As shown in Table~\ref{tab:reflect-interval}, we evaluate the sensitivity to this parameter across three games. Across all three games, $N=5$ and $N=2$ perform well, suggesting that moderate reflection intervals strike the right balance between map accuracy and stability. $N=1$ consistently underperforms: reflecting after every single episode provides too little new experience per cycle, leading to unstable map updates. Very large $N$ also degrades performance, as infrequent updates cause the map to lag behind the agent's accumulated experience.

\noindent\textbf{LLM Backbone.}
APEX consistently outperforms all baselines across three backbone LLMs, showing that its gains are not tied to any particular model. Full results are in Appendix~\ref{sec:backbone-results}.

\subsection{Cost Analysis}
\label{sec:cost}

Table~\ref{tab:cost} reports the API cost for each method on WebArena (812 tasks, 5 episodes per task). APEX's cost is comparable to other methods despite running additional reflection components, because milestone-guided prompts remain concise throughout training and naturally bound context growth.

\vspace{-1mm}

\section{Conclusion}
\vspace{-2mm}
\label{sec:conclusion}

We showed that self-evolving LLM agents often suffer from exploration collapse, converging to narrow behavioral patterns and missing superior strategies. APEX addresses the root cause: without an explicit strategy space, agents have no way to know what they have not tried. The strategy map provides this structure, Fork Discovery expands it, and Policy Selection navigates it. The result is consistent improvement over all baselines across nine Jericho games and WebArena, with each component contributing independently.

\noindent\textbf{Limitations.}
The strategy map's quality is bounded by the LLM's ability to identify milestones and infer unexplored directions, subtle or deeply hidden game mechanics may be overlooked by Fork Discovery. Moreover, APEX explore directions grounded in observed trajectories, and it cannot discover strategies that lie entirely outside the agent's past knowledge. Our evaluation focuses on text adventure games and web interaction tasks where progress naturally decomposes into discrete milestones. Tasks with continuous or less structured progress signals may require adapting how milestones are defined and evaluated.

\newpage
\bibliographystyle{unsrtnat}
\bibliography{references}

\newpage
\appendix

\section{Complete Workflow}
\label{sec:workflow}

Algorithm~\ref{alg:workflow} presents the complete APEX workflow. Here $\tilde{h}_t$ denotes the LLM-generated structured summary of episode $t$, and $\tilde{H}_t = \{\tilde{h}_i \mid t-N+1 \leq i \leq t\}$ denotes the summaries of the most recent $N$ episodes used by the reflection operations.

\renewcommand{\algorithmiccomment}[1]{\hfill // #1}
\begin{algorithm}[h]
\caption{APEX: Core Workflow}
\label{alg:workflow}
\begin{algorithmic}[1]
\REQUIRE Strategy map $\mathcal{G} = (V, E, \theta)$, episodes $T$, reflect interval $N$, discount $\gamma$
\FOR{$t = 1$ to $T$}
    \STATE $s \leftarrow \{v_0\}$, \quad $V_a \leftarrow \emptyset$, \quad $h_t \leftarrow []$
    \WHILE{episode not ended}
        \STATE $\mathcal{A}(s) \leftarrow \{v \notin s : \{u \mid (u,v) \in E\} \subseteq s\}$
        \STATE $v^* \leftarrow \argmax_{v \in \mathcal{A}(s)}\, f\bigl(n(v), \bar{r}(v), \text{Var}(v)\bigr)$ \COMMENT{score via Thompson Sampling}
        \STATE Execute $v^*$; observe $(o, r(v^*))$; \; $h_t \leftarrow h_t \cup \{(v^*, o, r(v^*))\}$; \; $V_a \leftarrow V_a \cup \{v^*\}$
        \IF{$v^*$ succeeded}
            \STATE $s \leftarrow s \cup \{v^*\}$
        \ENDIF
    \ENDWHILE
    \STATE $\tilde{h}_t \leftarrow \mathrm{LLM}(h_t)$; \quad $\tilde{H}_t \leftarrow \{\tilde{h}_{t-N+1}, \ldots, \tilde{h}_t\}$
    \IF{$t \bmod N = 0$}
        \STATE $(\Delta V^M, \Delta E^M) \leftarrow \mathrm{LLM}(\mathcal{G}_t,\, \tilde{H}_t)$ \COMMENT{Map Refinement}
        \STATE $\mathcal{G}'_t \leftarrow (V_t \cup \Delta V^M,\; E_t \cup \Delta E^M,\; \theta_t)$
        \FOR{each episode $i \in \{t-N+1,\ldots,t\}$, $v \in V_a^i$ in reverse topological order}
            \STATE $G_v \leftarrow r(v) + \gamma \sum_{u \in \mathrm{succ}(v) \cap V_a^i} G_u$ \COMMENT{Return Propagation}
            \STATE $n(v) \mathrel{+}= 1$; \quad $\bar{r}(v) \leftarrow \bar{r}(v) + \tfrac{1}{n(v)}\bigl(G_v - \bar{r}(v)\bigr)$
        \ENDFOR
        \STATE $(\Delta V^F, \Delta E^F) \leftarrow \mathrm{LLM}(\mathcal{G}'_t,\, \tilde{H}_t)$ \COMMENT{Fork Discovery}
        \STATE $\mathcal{G}_{t+1} \leftarrow (V_t \cup \Delta V^M \cup \Delta V^F,\; E_t \cup \Delta E^M \cup \Delta E^F,\; \theta_{t+1})$
    \ENDIF
\ENDFOR
\end{algorithmic}
\end{algorithm}

\section{Baseline Descriptions}
\label{sec:baselines}

We describe the five baselines used in our experiments.

\noindent\textbf{Static.}
The Static baseline is a memoryless agent that begins each episode from scratch with no knowledge of previous runs. It receives only the current episode's observations and a fixed system prompt, with no reflection, no accumulated memory, and no cross-episode information of any kind. Static serves as a lower bound: its occasional successes arise purely from behavioral diversity across independent episodes, not from any learning signal. Because it never exploits past experience, it cannot improve over time---but for the same reason, it never collapses into a narrow behavioral routine.

\noindent\textbf{Memory.}
The Memory baseline is an in-context learning agent that accumulates raw episode transcripts across runs. After each episode, the full trajectory (observations, actions, rewards) is appended to a growing context buffer. When the buffer exceeds the model's context window, a FIFO eviction policy drops the oldest episodes. At each step, the agent receives this accumulated history alongside the current observation, and selects actions informed by past experience. Memory improves over Static by giving the agent access to concrete prior examples, but it does not distill or structure that experience: the agent must extract useful patterns from raw transcripts, and as context grows, earlier episodes are lost.

\noindent\textbf{Reflexion~\cite{reflexion}.}
Reflexion augments the agent with a verbal self-reflection step after each episode. A second LLM call (or a second pass of the same model) reads the completed episode trajectory and produces a short structured reflection---typically a paragraph summarizing what went wrong, what should be tried differently, and what strategies were effective. These reflections are prepended to the system prompt of subsequent episodes, giving the agent a compact, curated summary of its own prior attempts. Reflexion is able to identify and fix local mistakes (e.g., avoiding a specific trap or correcting a failed action sequence), but its reflections describe what happened, not what the agent has not yet tried---leaving the unexplored strategy space implicit and untracked.

\noindent\textbf{EvoTest~\cite{evotest}.}
EvoTest frames agent improvement as an evolutionary optimization over the agent's configuration: its system prompt, in-context memory bank, and tool definitions are all treated as mutable components subject to iterative revision. After a batch of episodes, an LLM proposes mutations---additions, deletions, and edits---to each component. Mutations that improve performance are retained; others are discarded. This gradient-free evolution enables EvoTest to discover improvements that fixed reflection prompts may miss, including changes to tool behavior and memory structure. However, EvoTest does not maintain an explicit strategy map: its evolved components encode aggregate experience but provide no structured view of which strategies have been tried and which remain unexplored.

\noindent\textbf{ACE~\cite{ACE}.}
ACE (Agentic Context Engineering) treats the agent's context as an evolving playbook that accumulates and refines strategies through a modular grow-and-refine loop. After each episode, new strategies are extracted and appended to the playbook (the \emph{grow} step); periodically, an LLM consolidates redundant or outdated entries (the \emph{refine} step). The playbook is injected into the agent's context at each episode, providing a curated collection of reusable strategies. ACE differs from Reflexion in that it maintains a persistent, editable strategy bank rather than episode-level reflections, and from EvoTest in that its representation is human-readable and modular. Like both, however, ACE does not track which strategies remain unexplored: growth is driven by what the agent has encountered, not by what it has systematically avoided.

\section{Implementation Details}
\label{sec:impl-details}

\noindent\textbf{Hyperparameters.}
Table~\ref{tab:hyperparams} summarizes the hyperparameters used in all main experiments unless otherwise stated.

\begin{table}[h]
\centering
\small
\caption{APEX hyperparameters used in main experiments.}
\label{tab:hyperparams}
\begin{tabular}{@{}lll@{}}
\toprule
\textbf{Parameter} & \textbf{Value} & \textbf{Description} \\
\midrule
Reflect interval $N$ & 5 & Episodes between reflection cycles \\
Discount factor $\gamma$ & 0.6 & Return propagation discount \\
UCB constant $c$ & 10 & Exploration bonus coefficient \\
$\epsilon$ & 0.1 & Exploration probability ($\epsilon$-Greedy) \\
Max lessons & 20 & Global lesson buffer capacity \\
Fork Discovery ops & 6 & Max new milestones per reflection cycle \\
Exploration freeze & 30 & Episode at which Fork Discovery is disabled \\
Episodes & 50 & Per game (Jericho); 5 per task (WebArena) \\
Steps per episode & 120 & Environment step limit \\
\bottomrule
\end{tabular}
\end{table}

\noindent\textbf{Milestone Completion and Auto-Skip.}
At each step, the agent includes a \texttt{current\_milestone\_completed} flag in its structured output, by which it self-reports whether the current milestone has been achieved. The agent tracks how many steps it has spent on the current milestone; if this exceeds a patience threshold (40 steps for unvisited milestones with $n(v)=0$, 20 for previously attempted ones) without the agent setting this flag, the milestone is automatically skipped and the agent advances to the next one. To prevent immediate re-selection, the skipped milestone is added to a per-episode completed set, excluding it from Policy Selection for the remainder of the episode. This prevents the agent from wasting an entire episode stuck on a single milestone.

\noindent\textbf{Attempt Notes.}
Every $N$ episodes, as part of the reflection cycle, the system extracts attempt notes from $\tilde{H}_t$ for each milestone in the recent episode paths. Each note captures whether the milestone was achieved (with reward and step number) or failed (with failure reason), and is stored in a per-milestone rolling buffer of the 5 most recent attempts. These notes serve as the input to Stuck Node Diagnosis.

\noindent\textbf{Statistics During Deduplication.}
When Map Refinement prunes a duplicate node, the pruned node's bandit statistics $\theta_v$ and attempt notes are discarded; the surviving node retains its own accumulated statistics unchanged. This is justified because pruned nodes are typically lower-quality duplicates with inaccurate key actions or descriptions, making their statistics unrepresentative of the milestone's true value. Merging such statistics into the surviving node could corrupt its estimates.

\noindent\textbf{Stuck Node Diagnosis.}
Every $N$ episodes, the system identifies milestones that have been repeatedly attempted without success ($n(v) \geq 3$, $\bar{r}(v) \leq 0$), i.e., nodes that Policy Selection (Section~\ref{sec:selection}) keeps targeting but that the agent consistently fails to achieve. A diagnostic LLM call examines the attempt note buffer and the node's dependency context to identify the root cause (e.g., a missing prerequisite not yet encoded in $\mathcal{G}$) and produce targeted guidance. This guidance is injected into the agent's prompt when the milestone is next attempted. To avoid redundant calls, each node has a cooldown of $2N$ episodes before it can be diagnosed again.

\noindent\textbf{Global Lesson Extraction.}
Every $N$ episodes, after Fork Discovery, an LLM distills cross-episode patterns from $\tilde{H}_t$ into reusable categorical lessons. Three categories are maintained: \textsc{Penalty} lessons record actions that caused score loss, \textsc{Navigation} lessons capture movement constraints and shortcuts, and \textsc{Mechanic} lessons encode environment rules and object interactions. At most 5 new lessons are added per cycle. The full lesson buffer is injected into the system prompt at every step, giving the agent persistent cross-episode knowledge that complements the per-milestone guidance from the strategy map.

\noindent\textbf{Policy Selection Parameters.}
All three selection strategies share the same treatment of unvisited milestones: nodes with $n(v) = 0$ receive $+\infty$ scores and are selected uniformly at random, ensuring full coverage before exploitation begins.

\textbf{Thompson Sampling.} For milestones visited exactly once, we sample from $\mathcal{N}(\bar{r}(v), \sigma_{\text{prior}})$ with $\sigma_{\text{prior}} = 100$ to maintain high initial uncertainty. For milestones with multiple visits, the standard error is used with a minimum floor of $\sigma_{\min} = 1.0$ to prevent premature convergence.

\textbf{UCB.} The exploration constant is set to $c = 10$, tuned to match the reward scale of the games (per-episode scores typically range from 0 to 50).

\textbf{$\epsilon$-Greedy.} The exploration probability is set to $\epsilon = 0.1$.

\noindent\textbf{Strategy Map Grouping for WebArena.}
Unlike Jericho, where each game has a dedicated strategy map, WebArena comprises many distinct tasks. We group tasks dynamically: when a new task arrives, an LLM judges whether it is sufficiently similar to any previously seen task; if so, the new task reuses that task's strategy map, otherwise a new map is created. This allows the strategy map to accumulate knowledge across related tasks without requiring predefined task categories.

\noindent\textbf{Compute Resources.}
All experiments use API-based LLM inference with no local GPU required. A standard laptop suffices for orchestration; the only resource constraint is LLM API quota. No preliminary or failed experiments outside those reported were conducted at scale.

\section{Additional Results}
\label{sec:additional-results}

\subsection{Results with Standard Deviation}
\label{sec:final5-std}

Tables~\ref{tab:final5-std} and~\ref{tab:webarena-final3-std} report mean $\pm$ std for Jericho and WebArena respectively. All std values use sample standard deviation ($n-1$).

\begin{table}[h]
\centering
\caption{Final-5 mean $\pm$ std across all methods and nine Jericho games (50 episodes, 120 steps).}
\label{tab:final5-std}
\resizebox{\textwidth}{!}{%
\begin{tabular}{@{}lrrrrrrrrr@{}}
\toprule
\textbf{Method} & \textbf{Zork1} & \textbf{Balances} & \textbf{Zork3} & \textbf{Temple} & \textbf{Pentari} & \textbf{Ludicorp} & \textbf{Deephome} & \textbf{Detective} & \textbf{Ztuu} \\
\midrule
Static     & $39.8{\pm}5.3$  & $11.0{\pm}4.2$  & $2.8{\pm}0.4$ & $6.2{\pm}1.6$  & $6.0{\pm}2.2$  & $13.6{\pm}1.5$  & $17.2{\pm}1.8$   & $70.0{\pm}54.8$  & $0.0{\pm}0.0$ \\
Memory     & $46.0{\pm}13.9$ & $13.0{\pm}4.5$  & $2.4{\pm}0.5$ & $6.2{\pm}1.6$  & $7.0{\pm}2.7$  & $17.0{\pm}6.2$  & $29.0{\pm}21.1$  & $130.0{\pm}0.0$  & $0.0{\pm}0.0$ \\
Reflexion  & $59.8{\pm}6.7$  & $20.0{\pm}5.0$  & $3.0{\pm}0.0$ & $9.0{\pm}2.2$  & $10.0{\pm}0.0$ & $35.0{\pm}11.7$ & $29.6{\pm}6.7$   & $144.0{\pm}98.4$ & $5.0{\pm}0.0$ \\
EvoTest    & $42.0{\pm}4.5$  & $18.0{\pm}2.7$  & $2.4{\pm}0.5$ & $10.0{\pm}0.0$ & $16.0{\pm}5.5$ & $48.0{\pm}11.3$ & $56.8{\pm}31.3$  & $314.0{\pm}5.5$  & $25.0{\pm}2.7$ \\
ACE        & $46.0{\pm}9.6$  & $18.0{\pm}4.5$  & $2.2{\pm}0.4$ & $7.2{\pm}2.2$  & $19.0{\pm}2.2$ & $19.2{\pm}2.2$  & $8.4{\pm}1.5$    & $198.0{\pm}55.0$ & $3.4{\pm}5.3$ \\
APEX (Ours)& $\mathbf{73.0{\pm}13.7}$ & $\mathbf{23.0{\pm}2.7}$ & $\mathbf{4.6{\pm}0.9}$ & $\mathbf{20.8{\pm}1.1}$ & $\mathbf{46.0{\pm}4.2}$ & $\mathbf{58.8{\pm}9.1}$ & $\mathbf{103.4{\pm}0.9}$ & $\mathbf{322.0{\pm}4.5}$ & $\mathbf{29.0{\pm}6.7}$ \\
\bottomrule
\end{tabular}%
}
\end{table}

\begin{table}[h]
\centering
\caption{Final-3 mean $\pm$ std across all methods and five WebArena domains (5 episodes per task).}
\label{tab:webarena-final3-std}
\resizebox{\textwidth}{!}{%
\begin{tabular}{@{}lrrrrrrr@{}}
\toprule
\textbf{Method} & \textbf{Shopping} & \textbf{Reddit} & \textbf{GitLab} & \textbf{CMS} & \textbf{Map} & \textbf{Overall} \\
\midrule
Static      & $34.7{\pm}3.0$ & $45.7{\pm}0.4$ & $36.3{\pm}0.5$ & $44.8{\pm}0.8$ & $56.3{\pm}3.4$ & $42.3{\pm}0.2$ \\
Memory      & $39.1{\pm}1.7$ & $53.8{\pm}0.9$ & $53.9{\pm}1.8$ & $53.2{\pm}2.2$ & $55.2{\pm}2.5$ & $50.5{\pm}0.7$ \\
Reflexion   & $34.1{\pm}2.1$ & $36.2{\pm}2.0$ & $50.8{\pm}0.7$ & $46.7{\pm}0.0$ & $54.2{\pm}2.0$ & $44.3{\pm}0.8$ \\
EvoTest     & $38.9{\pm}1.7$ & $50.9{\pm}2.7$ & $54.4{\pm}0.0$ & $53.8{\pm}0.8$ & $52.9{\pm}3.2$ & $49.9{\pm}1.2$ \\
ACE         & $35.9{\pm}2.2$ & $52.5{\pm}1.3$ & $51.6{\pm}1.2$ & $52.4{\pm}1.4$ & $55.7{\pm}0.7$ & $48.6{\pm}1.4$ \\
\textbf{APEX (Ours)} & $\mathbf{42.9{\pm}2.0}$ & $\mathbf{68.2{\pm}0.8}$ & $\mathbf{60.2{\pm}0.6}$ & $\mathbf{55.1{\pm}1.9}$ & $\mathbf{57.6{\pm}1.3}$ & $\mathbf{55.9{\pm}0.5}$ \\
\bottomrule
\end{tabular}%
}
\end{table}

\subsection{LLM Backbone Comparison}
\label{sec:backbone-results}

To evaluate whether APEX's performance depends on the backbone LLM, we test three models while keeping all other settings identical. As shown in Table~\ref{tab:backbone}, APEX consistently outperforms all baselines across all three backbones, confirming that its gains reflect the strategy map and Fork Discovery rather than any particular model's capability.

\begin{table}[h]
\centering
\caption{LLM backbone comparison (Final-5).}
\label{tab:backbone}
\small
\setlength{\tabcolsep}{3pt}
\begin{tabular}{@{}l rrrrrr@{}}
\toprule
\textbf{Game} & Static & Memory & Reflexion & EvoTest & ACE & \textbf{APEX} \\
\midrule
\multicolumn{7}{c}{\textit{Gemini 3 Flash}} \\
\midrule
Zork1   & 39.8 & 46.0 & 59.8 & 42.0 & 46.0 & \textbf{73.0} \\
Pentari & 6.0  & 7.0  & 10.0 & 16.0 & 19.0 & \textbf{46.0} \\
Ztuu    & 0.0  & 0.0  & 5.0  & 25.0 & 3.4  & \textbf{29.0} \\
\midrule
\multicolumn{7}{c}{\textit{DeepSeek-V3.2}} \\
\midrule
Zork1   & 30.0 & 15.8 & 37.8 & 50.8 & 43.3 & \textbf{64.4} \\
Pentari & 16.0 & 5.0  & 22.0 & 26.0 & 10.0 & \textbf{43.0} \\
Ztuu    & 0.0  & 0.0  & 3.8  & 18.8 & 1.5  & \textbf{29.0} \\
\midrule
\multicolumn{7}{c}{\textit{GPT-5-mini}} \\
\midrule
Zork1   & 35.0 & 36.2 & 39.0 & 43.6 & 37.4 & \textbf{59.8} \\
Pentari & 5.0  & 10.0 & 5.0  & 11.0 & 8.0  & \textbf{42.0} \\
Ztuu    & 5.0  & 2.8  & 3.4  & 8.4  & 4.8  & \textbf{19.2} \\
\bottomrule
\end{tabular}
\end{table}

\subsection{Full Learning Curves}
\label{sec:full-curves}

\begin{figure}[h]
  \centering
  \includegraphics[width=\linewidth]{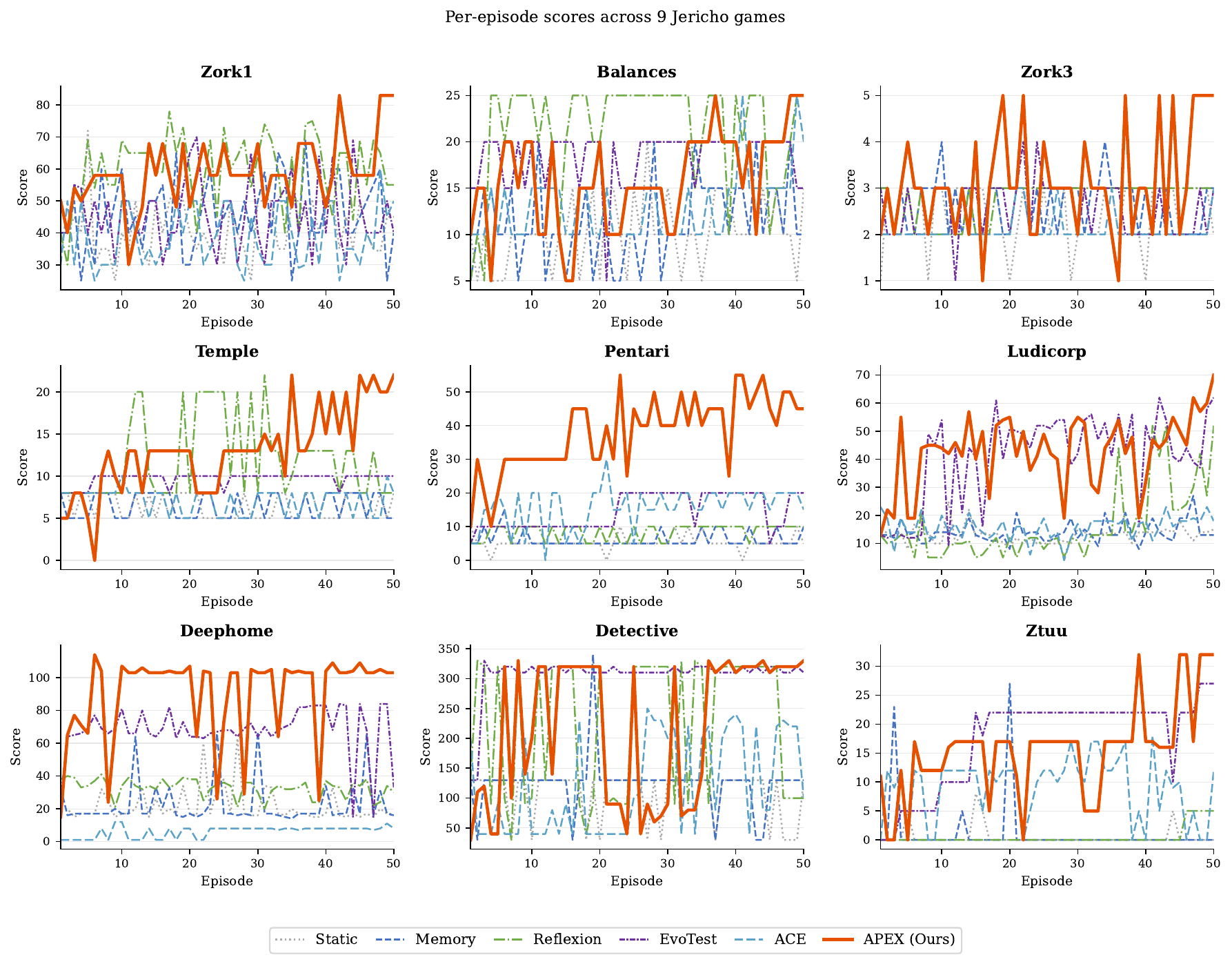}
  \caption{Per-episode scores across 50 episodes on all nine Jericho games.}
  \label{fig:learning-curves-full}
\end{figure}

The learning curves on nine Jericho games are shown in Figure~\ref{fig:learning-curves-full}.

\subsection{Prompt Templates}
\label{sec:prompts}

This section presents the core prompt templates used in APEX. Dynamic content (episode data, map state, etc.) is injected at the positions marked with placeholders.

\subsection{Action Prompt}
\label{sec:prompt-action}

The agent receives a system prompt and a user prompt at each step. The system prompt is composed of a base instruction, global lessons, and strategy guidance derived from the current milestone.

\begin{promptbox}[Action Prompt --- System Prompt]
\small
You are an expert agent aiming to complete an interactive task. Rewards are given for making progress. Select promising actions based on the current state and memory of past interactions.

LESSONS FROM PAST EXPERIENCE (follow these strictly):

- [PENALTY] \textit{\{lesson text\}}

- [NAVIGATION] \textit{\{lesson text\}}

- [MECHANIC] \textit{\{lesson text\}}
\end{promptbox}

\begin{promptbox}[Action Prompt --- User Prompt]
\small
RECENT STEPS:

Step \textit{i}: State: \textit{\{state\}} Action: \textit{\{action\}} Reward: \textit{\{reward\}}

$\ldots$

LAST 8 ACTIONS: \textit{action$_1$} $\to$ \textit{action$_2$} $\to$ $\ldots$

CURRENT STATE: \textit{\{current observation, agent state / inventory\}}

PROGRESS: Step \textit{t}/\textit{T} (\textit{remaining} steps left) | Reward: \textit{s}

CURRENT MILESTONE: \textit{\{milestone description\}}

\quad Key action sequence: \textit{\{key actions\}}

\quad Known pitfalls: At [\textit{\{location\}}]: agents often loop doing [\textit{\{pattern\}}]. Instead, try: [\textit{\{escape action\}}]

TASK:

1. Analyze your progress: What have you achieved? What's your next objective?

2. Propose your best next action with reasoning

RESPONSE FORMAT (JSON):

\{"progress\_analysis": "...", "current\_milestone\_completed": false, "next\_objective": "...", "reasoning": "...", "action": "..."\}

KEY RULES:

- If you haven't received rewards recently, you are likely stuck --- try a fundamentally different approach

- LOOP DETECTION: Look at the LAST ACTIONS trail. If cycling between the same states, do something completely different

\end{promptbox}

\subsection{Episode Summary Prompt}
\label{sec:prompt-summary}

After each episode, an LLM generates a milestone-centric structured summary used by all reflection modules. Milestones are significant task progress steps identified from the trajectory, regardless of whether they directly earned a reward.

\begin{promptbox}[Episode Summary --- System Prompt]
\small
You are an expert at analyzing agent episode trajectories. Your job is to extract milestones --- the key sub-goals that represent significant task progress --- and analyze the path to each one.
\end{promptbox}

\begin{promptbox}[Episode Summary --- User Prompt]
\small
Final score: \textit{s}, Total steps: \textit{T}

STRATEGY MAP: \textit{\{strategy map\}}

EPISODE TRAJECTORY:

Step \textit{i}: STATE: \textit{observation} | ACTION: \textit{action} | REWARD: +\textit{r}

$\ldots$

Write the summary in this format:

\begin{itemize}
\item \textbf{Milestones Achieved}: matched to known nodes or marked [NEW] if unrecognized
\item \textbf{Penalties Received}: actions that caused score loss
\item \textbf{Milestones NOT Achieved}: with failure reason and missing prerequisites
\item \textbf{Unexplored Opportunities}: actions available or observed but never tried (untried valid actions, unexamined objects, unexplored paths)
\end{itemize}

Critical: identify milestones by meaningful task progress steps, not by incidental state features.
\end{promptbox}

\subsection{Map Refinement Prompt}
\label{sec:prompt-tree-update}

Map Refinement updates the strategy map based on observed facts from recent episodes. The LLM receives the current map state and episode summaries, then proposes structural operations.

\begin{promptbox}[Map Refinement --- System Prompt]
\small
You are an expert at analyzing agent episode results and maintaining a strategy map.

The agent's knowledge is organized as a MILESTONE DAG (directed acyclic graph). Each node has a ``deps'' list meaning ``this milestone REQUIRES these specific prerequisites to be completed first''. deps=[] means the milestone is independent.

Your job is to UPDATE the DAG based on OBSERVED FACTS from the episodes below. You must ONLY encode things that actually happened --- do NOT speculate.

RULES:

(1) Every significant event that \textbf{actually occurred} should correspond to a node --- reward changes, state transitions, key resource acquisitions. Do NOT add nodes for things the agent never tried.

(2) Fix incorrect key\_actions with update\_node; do not rename nodes with confirmed rewards.

(3) Prune only to remove duplicate nodes.

(4) Only add a dependency when the milestone is impossible without the prerequisite.

(5) When updating an existing node, make only minor corrections (fix key\_actions, adjust dependency edges). If the core goal of the milestone changes fundamentally, create a new node with a fresh identity instead.

(6) To correct a wrong dependency edge, use \texttt{update\_deps} to move the node to a different parent. This preserves the node's statistics and subtree.
\end{promptbox}

\begin{promptbox}[Map Refinement --- User Prompt]
\small
STRATEGY MAP: \textit{\{strategy map\}}

EPISODE SUMMARIES: \textit{\{structured summaries\}}

Step 1 --- Analyze each episode: list every significant event and check against DAG.

Step 2 --- Analyze milestone failures: check for missing prerequisites.

Step 3 --- Dedup check and dependency audit.

Step 4 --- Propose DAG operations: add\_child, add\_branch, update\_node, update\_deps, or prune.
\end{promptbox}

\subsection{Fork Discovery Prompt}
\label{sec:prompt-dpm}

Fork Discovery identifies unexplored directions --- actions the agent observed or had available but never pursued. It takes episode summaries (Section~\ref{sec:prompt-summary}) and the current map as input.

\begin{promptbox}[Fork Discovery --- System Prompt]
\small
You are an expert at identifying unexplored opportunities in agent episode trajectories. Your task is to identify actions the agent observed or had available but never pursued, which could unlock new content.

CRITICAL CONSTRAINTS: Every proposed milestone must be a concrete, executable action grounded in evidence from the episodes. Limit to at most 6 new milestones per reflection.
\end{promptbox}

\begin{promptbox}[Fork Discovery --- User Prompt]
\small
STRATEGY MAP: \textit{\{strategy map\}}

EPISODE TRAJECTORIES: \textit{\{episode summaries\}}

Identify unexplored opportunities: find actions or directions that were available or observable in the episodes but never attempted. Propose new milestone nodes with appropriate dependency edges to encode these discoveries into the strategy map.
\end{promptbox}

\subsection{Stuck Node Diagnosis Prompt}
\label{sec:prompt-diagnosis}

Stuck Node Diagnosis analyzes milestones that persistently fail ($n \geq 3$, $\bar{r} \leq 0$) to produce targeted guidance.

\begin{promptbox}[Stuck Node Diagnosis --- System Prompt]
\small
You are an expert at diagnosing why an agent keeps failing at certain milestones.

Analyze the stuck node's attempt notes and context to determine: (1) WHY the agent keeps failing (root cause), (2) What specific action the agent should take next time, (3) Whether the failure is due to a missing prerequisite or unmet dependency.

\end{promptbox}

\begin{promptbox}[Stuck Node Diagnosis --- User Prompt]
\small
Analyze this stuck node:

--- \textit{node\_id}: \textit{milestone} ---

\quad key\_action: \textit{\{actions\}}

\quad visits: \textit{n}, avg\_reward: \textit{r}

\quad Recent attempt notes: \textit{\{last 5 attempts\}}

\quad Depends on: \textit{\{upstream nodes\}}

\quad Downstream: \textit{\{dependent nodes\}}

Respond with: \textbf{Root cause} (why it keeps failing), \textbf{Next action} (what to try), \textbf{Missing prerequisite} (if applicable).
\end{promptbox}

\subsection{Global Lesson Extraction Prompt}
\label{sec:prompt-lessons}

Global Lesson Extraction distills cross-episode patterns into reusable categorical lessons.

\begin{promptbox}[Global Lesson Extraction --- System Prompt]
\small
You are an expert at extracting reusable lessons from agent episode trajectories.

Extract GENERAL lessons the agent should remember across all future episodes. Three categories: PENALTY records actions that caused reward loss; NAVIGATION captures movement constraints and shortcuts; MECHANIC encodes environment mechanics and object interactions.

Accuracy rules: only record observed facts, require evidence for conditional discoveries. Maximum 5 new lessons per call.
\end{promptbox}

\begin{promptbox}[Global Lesson Extraction --- User Prompt]
\small
\textit{\{Recent episode summaries\}}

EXISTING LESSONS: \textit{\{current lesson buffer\}}

What general lessons should the agent remember for all future episodes?
\end{promptbox}

\section{Asset Licenses}
\label{sec:licenses}

Table~\ref{tab:licenses} lists the open-source assets used in this paper along with their licenses.

\begin{table}[h]
\centering
\small
\caption{Licenses for assets used in this paper.}
\label{tab:licenses}
\begin{tabular}{@{}lll@{}}
\toprule
\textbf{Asset} & \textbf{Source} & \textbf{License} \\
\midrule
Jericho~\cite{jerichogame} & \url{https://github.com/microsoft/jericho} & MIT \\
WebArena~\cite{webarena} & \url{https://github.com/web-arena-x/webarena} & Apache 2.0 \\
Reflexion~\cite{reflexion} & \url{https://github.com/noahshinn/reflexion} & MIT \\
EvoTest~\cite{evotest} & \url{https://github.com/microsoft/EvoAgent} & MIT \\
ACE~\cite{ACE} & \url{https://github.com/acewrench/ACE} & MIT \\
\bottomrule
\end{tabular}
\end{table}

\end{document}